%% file: root.tex
\documentclass[letterpaper, 10 pt, conference]{ieeeconf}  % Comment this line out if you need a4paper

\IEEEoverridecommandlockouts                              % This command is only needed if 
\usepackage{graphics} % for pdf, bitmapped graphics files
\usepackage{amsmath} % assumes amsmath package installed
\usepackage{amssymb}  % assumes amsmath package installed
\usepackage{algorithm,algorithmic}
\usepackage{graphicx}
\usepackage{hyperref}
\usepackage{booktabs}
\usepackage{flushend}
\usepackage{xcolor}

\title{\LARGE \bf
A Few Shot Adaptation of Visual Navigation Skills \\ to New Observations using Meta-Learning
}

\author{
Qian Luo$^{1}$, Maks Sorokin$^{1}$, Sehoon Ha$^{12}$ % please keep Maks as my first name :) 
\thanks{$^{1}$ Georgia Institute of Technology, Atlanta, GA, 30308, USA}
\thanks{$^{2}$ Robotics at Google, Mountain View, CA, 94043, USA }
\thanks{Emails: {\tt\small \{luoqian,maks,sehoonha\}@gatech.edu}}
}

\begin{document}

\maketitle
\thispagestyle{empty}
\pagestyle{empty}

\input{defs}

%%%%%%%%%%%%%%%%%%%%%%%%%%%%%%%%%%%%%%%%%%%%%%%%%%%%%%%%%%%%%%%%%%%%%%%%%%%%%%%%
\begin{abstract}

Target-driven visual navigation is a challenging problem that requires a robot to find the goal using only visual inputs.
Many researchers have demonstrated promising results using deep reinforcement learning (deep RL) on various robotic platforms, but typical end-to-end learning is known for its poor extrapolation capability to new scenarios.
Therefore, learning a navigation policy for a new robot with a new sensor configuration or a new target still remains a challenging problem.
In this paper, we introduce a learning algorithm that enables rapid adaptation to new sensor configurations or target objects with a few shots.
We design a policy architecture with latent features between perception and inference networks and quickly adapt the perception network via meta-learning while freezing the inference network.
Our experiments show that our algorithm adapts the learned navigation policy with only three shots for unseen situations with different sensor configurations or different target colors.
We also analyze the proposed algorithm by investigating various hyperparameters.

\end{abstract}

%%%%%%%%%%%%%%%%%%%%%%%%%%%%%%%%%%%%%%%%%%%%%%%%%%%%%%%%%%%%%%%%%%%%%%%%%%%%%%%%

\input{Introduction}
\input{Related}

\input{Method}

\input{Experiment}
\input{Conclusion}

\bibliography{ref}
\bibliographystyle{IEEEtran}
\end{document}

%% file: defs.tex
%% editing comment

\newcommand{\qian}[1]{\textcolor{blue}{{Qian: #1}}}
\newcommand{\sehoon}[1]{\textcolor{magenta}{{Sehoon: #1}}}
\newcommand{\maks}[1]{\textcolor{red}{{Maks: #1}}}
\newcommand{\revised}[1]{\textcolor{blue}{#1}}

\newcommand{\original}[1]{\textcolor{magenta}{Original: #1}}
\newcommand{\eqnref}[1]{Equation~(\ref{eq:#1})}
\newcommand{\figref}[1]{Figure~\ref{fig:#1}}
\newcommand{\algref}[1]{Algorithm~\ref{alg:#1}}
\newcommand{\tabref}[1]{Table~\ref{tab:#1}}
\newcommand{\secref}[1]{Section~\ref{sec:#1}}

%% ignore text
\long\def\ignorethis#1{}
\newcommand{\myparagraph}[1]{\noindent\textbf{{#1}}}

%% abbreviations
\newcommand{\etal}{{\em{et~al.}\ }}
\newcommand{\eg}{e.g.\ }
\newcommand{\ie}{i.e.\ }

%% reference shortcuts
\newcommand{\figtodo}[1]{\framebox[0.8\columnwidth]{\rule{0pt}{1in}#1}}

%\renewcommand{\eqref}[1]{Equation~(\ref{eq:#1})}

%% frequently used mathematical structures

\newcommand{\pdd}[3]{\ensuremath{\frac{\partial^2{#1}}{\partial{#2}\,\partial{#3}}}}

%% New commands for Sehoon!
\newcommand{\mat}[1]{\ensuremath{\mathbf{#1}}}
\newcommand{\set}[1]{\ensuremath{\mathcal{#1}}}

% math macros
\newcommand{\vc}[1]{\ensuremath{\mathbf{#1}}}
\newcommand{\vEndEff}{\ensuremath{\vc{d}}}
\newcommand{\vRelMove}{\ensuremath{\vc{r}}}
\newcommand{\sSet}{\ensuremath{S}}

\newcommand{\vControl}{\ensuremath{\vc{u}}}
\newcommand{\vPoint}{\ensuremath{\vc{p}}}
\newcommand{\sSpringCoef}{{\ensuremath{k_{s}}}}
\newcommand{\sDamperCoef}{{\ensuremath{k_{d}}}}
\newcommand{\vHandle}{\ensuremath{\vc{h}}}
\newcommand{\vForce}{\ensuremath{\vc{f}}}

\newcommand{\mTransChain}{\ensuremath{\vc{W}}}
\newcommand{\mRotateTrans}{\ensuremath{\vc{R}}}
\newcommand{\sJoint}{\ensuremath{q}}
\newcommand{\vJoint}{\ensuremath{\vc{q}}}
\newcommand{\mJoint}{\ensuremath{\vc{Q}}}
\newcommand{\mMass}{\ensuremath{\vc{M}}}
\newcommand{\sMass}{\ensuremath{{m}}}
\newcommand{\vGravity}{\ensuremath{\vc{g}}}
\newcommand{\vConstr}{\ensuremath{\vc{C}}}
\newcommand{\sConstr}{\ensuremath{C}}
\newcommand{\vCOM}{\ensuremath{\vc{x}}}
\newcommand{\sGeneralForce}[1]{\ensuremath{Q_{#1}}}
\newcommand{\vStateVar}{\ensuremath{\vc{y}}}
\newcommand{\vControlVar}{\ensuremath{\vc{u}}}
\newcommand{\tr}[1]{\ensuremath{\mathrm{tr}\left(#1\right)}}

%%%%%%%%%%%%%%%%%%%%%%%%%%%%%%%%%%%%%%%%%%%%%%%%%%%%%%%%%%%%%%%%%%%
%
% Here are a bunch of macros, mostly for math.
%
%%%%%%%%%%%%%%%%%%%%%%%%%%%%%%%%%%%%%%%%%%%%%%%%%%%%%%%%%%%%%%%%%%%

\renewcommand{\choose}[2]{\ensuremath{\left(\begin{array}{c} #1 \\ #2 \end{array} \right )}}

\newcommand{\gauss}[3]{\ensuremath{\mathcal{N}(#1 | #2 ; #3)}}

\newcommand{\pctab}{\hspace{0.2in}}
\newenvironment{pseudocode} {\begin{center} \begin{minipage}{\textwidth}
                             \normalsize \vspace{-2\baselineskip} \begin{tabbing}
                             \pctab \= \pctab \= \pctab \= \pctab \=
                             \pctab \= \pctab \= \pctab \= \pctab \= \\}
                            {\end{tabbing} \vspace{-2\baselineskip}
                             \end{minipage} \end{center}}
\newenvironment{items}      {\begin{list}{$\bullet$}
                              {\setlength{\partopsep}{\parskip}
                                \setlength{\parsep}{\parskip}
                                \setlength{\topsep}{0pt}
                                \setlength{\itemsep}{0pt}
                                \settowidth{\labelwidth}{$\bullet$}
                                \setlength{\labelsep}{1ex}
                                \setlength{\leftmargin}{\labelwidth}
                                \addtolength{\leftmargin}{\labelsep}
                                }
                              }
                            {\end{list}}
\newcommand{\newfun}[3]{\noindent\vspace{0pt}\fbox{\begin{minipage}{3.3truein}\vspace{#1}~ {#3}~\vspace{12pt}\end{minipage}}\vspace{#2}}

\newcommand{\key}{\textbf}
\newcommand{\fun}{\textsc}

%\def\shortcite{\def\citename##1{}\@internalcite}

% Local Variables:
% TeX-master: "paper"
% End:

%% file: Introduction.tex
\section{Introduction} \label{sec:intro}

Visual navigation is a fundamental skill for robotic creatures to conduct real-world autonomous missions, such as logistics, surveillance, or home-assistance.
However, learning visual navigation policy is a challenging problem that involves multiple aspects of robotics research, from scene understanding to motion planning.
Recent advances in deep learning have enabled robotic agents to learn effective navigation policies in various scenarios~\cite{jung2018perception,chiang2019rl,amini2019variational}.
However, deep RL often requires a tremendous amount of computational costs.
For instance, the work of Wijmans et al.~\cite{wijmans2019dd} requires a large GPU cluster with $64$ nodes to solve a point-to-point navigation problem, which corresponds to six-months of GPU time.

\begin{figure}
    \centering
    \includegraphics[width=0.8\linewidth]{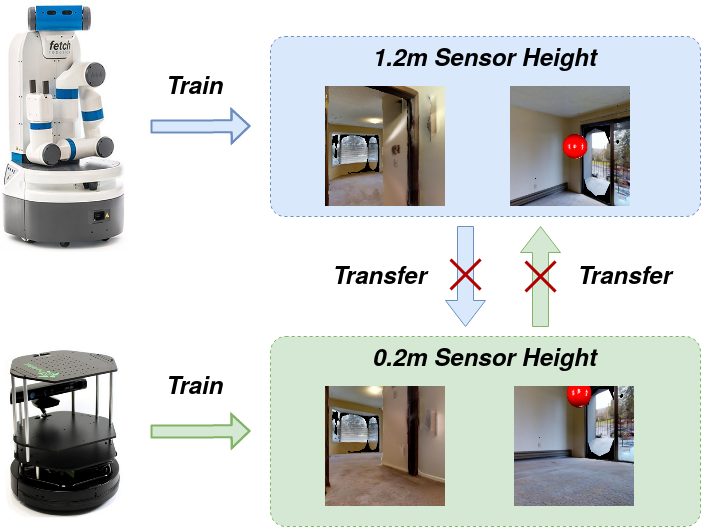}
    \caption{The visual features of robots of different sensor heights are quite different, making it hard to directly transfer the learned policy to another robot. Visual navigation for heterogeneous robots must resolve perspective differences.}
    \label{fig:perspective}
\end{figure}

Learning visual navigation becomes even more challenging if we start to consider a lot of heterogeneous robots needing navigation skills for their missions, such as wheeled robots for warehouse logistics, legged robots for wilderness monitoring, or drones for last-mile delivery.
A naive approach is to learn a navigation policy for each robot independently, but it will be computationally expensive.
To make matters worse, there is no guarantee that the same scheme will lead to the same successful policy on all the robots.
An alternative approach is to transfer a learned policy from one robot to another robot: however, heterogeneous robots have different sensing and actuation capabilities, making the transfer difficult.
Figure~\ref{fig:perspective} illustrates camera inputs of different robots(different sensor heights) that are sampled from the same location, which present noticeable different visual features.

In this work, we propose a novel algorithm for learning a visual navigation policy that can quickly adapt to new observations induced by different camera configurations or different target objects.
We assume the same action space across different robots; under this assumption, all agents are controlled using the same abstract navigation commands, such as go straight, turn left, turn right, and stop.
Under this assumption, the key technical challenges here are to extract invariant knowledge shared across different observation spaces and to adapt the learned policy to a new scenario with a minimal amount of additional data.

We tackle this problem based on two steps: the \emph{baseline agent learning} step and the \emph{adaptation} step.
During the baseline agent learning step, we train an effective navigation agent that consists of two components:
1) a perception network that encodes visual inputs into the latent variable and 2) an inference network that infers the cost-to-go value and optimal actions from the latent variable.
During the adaptation phase, we only adapt the perception network using supervised learning (SL) while freezing the inference network.
Because SL requires a lot of samples from the target environment that are not available for real-world applications, we propose a few-shot adaptation method via meta-learning.

We evaluate the proposed algorithm on a target-driven navigation problem on the Habitat simulator~\cite{habitat19iccv}. 
% We investigate $???$ scenarios: the changes of the camera height, the field of view, and the target object.
We investigate the adaptation of navigation skills to unseen scenarios with different camera heights and different target objects.
Our experimental results indicate that the learned agent with one visual observation space can be quickly adapted to a new scenario by using less than five shots, which is enabled by our novel method of meta-learning on the latent space.
We further validate the algorithm's design by comparing different hyper-parameters, the dimension of the latent parameters and the number of meta update steps.

%% file: Related.tex
\section{Related Work}
\label{sec:rel}

% We draw our inspiration from work done in the area of deep-reinforcement learning, target-driven visual navigation, and meta-reinforcement learning.

\subsection{DRL based Visual Navigation}
% Introduce recent research work solving target-driven navigation problem using Deep RL.
Over the past few years, a tremendous effort has been put towards enabling autonomous navigation agents in realistic environments. 
For this purpose, many environments with various fidelity have been developed.
The early work of Mirowski et al.~\cite{mirowski2016learning} provides game-like maze environments, while recent simulators like MINOS \cite{savva2017minos}, Habitat \cite{habitat19iccv}, and iGibson \cite{xiazamirhe2018gibsonenv, xia2020interactive}, provide a physics-enabled simulated environment with a photo-realistic rendering for training and evaluation of autonomous agents performing various tasks such as object manipulation, navigation, and search. 
These simulators enable the agents to learn the task via trial-and-errors, over millions or billions of environment interactions, aiming for successful sim-to-real transfer.

Recently, Decentralized Distributed Proximal Policy Optimization (DD-PPO)~\cite{wijmans2019dd} demonstrates that a near-perfect point-goal agent can be trained to navigate based on visual observations and GPS information in an end-to-end fashion.
% \qian{DD-PPO learns a policy in a large number of indoor environments when trained for six months of a single GPU time, which is equivalent to 80 years of human experience.
% % \qian{In the DD-PPO paper, the success rates of RGB and RGBD are both over 99\%. However, all their experiments are based on point-goal navigation, which is much easier than our object navigation problem. Personally, I prefer not to write too much about DD-PPO.(We could just mention it for a few lines) The paper we should focus on(write much about it) is \cite{zhu2017target}, which is the most important paper in target-driven navigation recent years, and is the baseline we try to study in the experiment part.}. 
% While DD-PPO learns a single navigation policy conditioned on the coordinate information about the target for all scenes, }
Yuke Zhu et al. \cite{zhu2017target} trains scene-specific expert policies while demonstrating how the control policy can be conditioned on visual target information, such as a picture of the goal location in addition to the agent's first-person view observation.

To enable longer-horizon navigation with spatio-temporal dependencies, the work of Fang~\cite{fang2019scene} proposes a novel network architecture using Transformer ~\cite{vaswani2017attention} for visual navigation, which shows better performance than both reactive and Long short-term memory (LSTM)~\cite{hochreiter1997long} agents. 
Active Neural SLAM \cite{chaplot2020learning, chaplot2020object} shows that the careful hierarchical decomposition of the navigation policy leads to reduced sample complexity compared to typical end-to-end agents.

We utilize the Habitat simulator \cite{habitat19iccv} and PPO~\cite{schulman2017proximal} algorithm to train a vision-based target-driven expert policy. 
Our agent perceives the RGB visual information of the environment that contains a single visual target goal.

\subsection{Skill Generalization in Visual Navigation}
% Introduce those research work focus on the generalization ability in visual navigation.

Various methods for generalizing the learned agents to new scenarios have been proposed \cite{zhu2017target, ye2018active, devo2020towardsgen, tobin2017domain, sax2018mid, mousavian2019visual, gordon2019splitnet}. 
One of the popular approaches to generalize over the various sensor and task configurations is Domain Randomization (DR)~\cite{tobin2017domain, sadeghi2016cad2rl}, which trains an agent in randomized environments. 
% However, one known drawback of DR is to learn a conservative policy that is suboptimal to many tasks.

% While training with DR the navigation policy is exposed to a wide distribution of sensor/task configurations, however, there is still a single policy that is expected cover/operate in all of the configurations which might be an unreasonable expectation. \maks{Qian: if we have DR results could you share them, so I can write more about the drawbacks here.}

An alternative approach to address the generalization problem is a hierarchical approach that decomposes the problem into multiple sub-components.
For instance, the object-search task can be decomposed into the detection of object-specific information and navigation based on the extracted information.
The work of Ye et al.~\cite{ye2018active} jointly trains an object recognition model for predicting the object mask and a control policy.
At the adaptation phase, only the object recognition model is adapted with supervised learning, while the control policy remains unchanged. 
Similarly, in the work of Devo et al.~\cite{devo2020towardsgen}, the architecture is decomposed into two networks: object recognition network and navigation/exploration, where only the recognition part is adapted at the transfer stage. 
Yuke Zhu et al. \cite{zhu2017target} studies the generalization of RL in different scenes and tasks by sharing feature embeddings of perception network across scene-specific expert policies. With such design the perception network is ought to produce generalized features that are useful in different scenes.

Our policy is similarly decomposed into two components; however, we do not rely on any assumption regarding the task at hand. The expert policy (prior to adaptation) is trained in an end-to-end fashion with a constrain of a bottleneck between perception and inference components. We keep the inference network fixed at an adaptation stage and only adapt the perception part. While the task selected for experiments is related to navigation, our method can be applied to a wide range of vision-based control problems.

\subsection{Meta-learning and Few Shot Adaptation}
% Introduce representative research work on meta learning, meta reinforcement learning...

As opposed to learning a single policy exhibiting robust behaviors in all scenarios, researchers~\cite{finn2017model, nichol2018first, yang2019norml, rakelly2019efficient, rusu2018meta, yin2019meta} have developed algorithms to quickly adapt the learned policy to a new scenario, which is referred to as meta-learning.
One notable algorithm is Model-agnostic Meta-learning (MAML)~\cite{finn2017model}, which is a meta-learning algorithm that can be applied to both supervised learning and reinforcement learning.
During the training phase of MAML, the optimizer brings the network parameters into the state where they are ready for fast adaptation to a new task.
By taking a few gradient steps during the meta-adaptation phase, the agent becomes ready for the testing domain. 
In the work of Wortsman et al.~\cite{wortsman2019learning}, a self-supervised interaction objective is combined with MAML to enable the navigation agent to adapt throughout testing rollouts.
Li et al. propose ~\cite{li2020unsupervised} to jointly meta-train a high-level policy with sub-skills in an unsupervised fashion and only adapt the high-level policy at the adaptation phase.
Our approach also utilizes the MAML for training the perception network without relying on the learning of the sub-skills.
We demonstrate that latent embedding enables rapid adaptation with only a handful of the target images set to the same configuration as in the source environment.

%% file: Method.tex
\section{A Few Shot Adaptation with Meta Learning} \label{sec:method}
% In this section, we introduce how we accomplish a few shot adaptation of navigation skills based on a trained agent. 
% Firstly we present the problem we hope to solve. Then we introduce our navigation baseline and 'latent space', which represents the general navigation skills. 
% We then explain how to use the learned latent space to enable fast adaptation to new tasks via limited amount of data, with supervised learning and meta learning.
In this section, we present our algorithm that adapts a learned visual navigation agent to a new observation function, which can be caused by different camera heights or different target objects.
Our key idea is to train a baseline navigation agent with two components; the perception and inference networks connected by \emph{latent features}, and to adapt only the perception network efficiently via meta-learning while freezing the inference network. 

The rest of the section is organized as follows.
We first describe the problem formulation (Section~\ref{sec:problem}) and explain the training of a baseline navigation agent (Section~\ref{sec:baseline}).
Then we describe fast adaptation with supervised learning (Section~\ref{sec:supervised}) and meta learning (Section~\ref{sec:meta}).

\subsection{Problem Formulation} \label{sec:problem}

% Reinforcement learning (RL) aims to learn a policy that maximizes
% the expected sum of rewards~\cite{sutton1998reinforcement}. 
% We formulate the visual navigation task as Partially Observable Markov Decision Processes (PoMDPs), $(\mathcal{S}, \mathcal{O}, \mathcal{A}, \mathcal{T}, r, p_0, \gamma)$, where $\mathcal{S}$ is the state space, $\mathcal{O}$ is the observation space, $\mathcal{A}$ is the action space, $\mathcal{T}$ is the transition function, $r$ is the reward function, $p_0$ is the initial state distribution and $\gamma$ is a discount factor. We take the approach of model-free reinforcement learning to find a policy $\pi$, such that it maximizes the accumulated reward:
% \begin{equation}
% \label{eq:rl}
%     J(\pi) = \mathbb{E}_{\mathbf{s}_0, \mathbf{a}_0, \dots, \mathbf{s}_T} \sum_{t=0}^{T} \gamma^t r(\mathbf{s}_t, \mathbf{a}_t),
% \end{equation}
% where $\mathbf{s}_0 \sim p_0$, $\mathbf{a}_t \sim \pi(\mathbf{o}_t)$, $\vc{o}_t \sim c(\vc{s}_t)$ and $\mathbf{s}_{t+1}=\mathcal{T}(\mathbf{s}_t, \mathbf{a}_t)$.
% We assume that different robots have their own observation function $c^i(\vc{s}_t)$, which is caused by sensor configurations, including camera types, positions, or field of views.
% However, we assume that all the robots share the same state space $\mathcal{S}$ and action space $\mathcal{A}$, which may not hold if we start to consider robots with different actuators, such as legged robots and wheeled robots.

% Sehoon: define observation, action, and reward more closely.

Let us consider a target-driven visual navigation problem, where the agent aims to learn a policy to find the given target with egocentric visual inputs. 
We formulate this task as Partially Observable Markov Decision Processes (PoMDPs), $(\mathcal{S}, \mathcal{O}, \mathcal{A}, \mathcal{T}, r, p_0, \gamma)$, where $\mathcal{S}$ is the state space, $\mathcal{O}$ is the observation space, $\mathcal{A}$ is the action space, $\mathcal{T}$ is the transition function, $r$ is the reward function, $p_0$ is the initial state distribution and $\gamma$ is a discount factor.

In a typical navigation episode, at each time $t$, the agent receives an observation $\vc{o}_t$ based on current state $\vc{s}_t$, $\vc{o}_t = c(\vc{s}_t)$.
After the agent takes an action $\mathbf{a}_t \sim \pi(\mathbf{o}_t)$ based on the observation $\vc{o}_t$, it gets reward $r$ from the environment and arrives at a new state $\mathbf{s}_{t+1}=\mathcal{T}(\mathbf{s}_t, \mathbf{a}_t)$.
The goal of reinforcement learning~\cite{sutton1998reinforcement} is to find the optimal policy that maximizes the accumulated reward:
\begin{equation}
\label{eq:rl}
    J(\pi) = \mathbb{E}_{\mathbf{s}_0, \mathbf{a}_0, \dots, \mathbf{s}_T} \sum_{t=0}^{T} \gamma^t r(\mathbf{s}_t, \mathbf{a}_t).
\end{equation}

% Define an observation space. Talk about the adaptation.
\myparagraph{Observation Space.}
In our problem, we assume that each robotic agent has its own observation function $\vc{o}_t = \tilde{c}(\vc{s}_t) \in \mathcal{\tilde{O}}$, which represents its unique camera perspective, possibly with the target object class present in observation.
Here, we regard $\vc{s}_t$ as the ground truth state at time $t$, which represents the agent's position and orientation.
For instance, Turtlebot \cite{turtlebot} and Fetch \cite{wise2016fetch} robots have different camera heights of $0.15-0.55$m and $1.1-1.5$m, which results in distinctive visual features in input images from the same location.
Also, the observed visual inputs of different targets from the same position will be different from each other.
We assume that an observation function $\tilde{c}$ typically returns a single RGB or RGB-D image.
This robot-specific observation function defines our unique challenge of learning a common navigation policy that can be generalized to a new observation space.

% Define an action space
\myparagraph{Action Space.}
From the given observation $\vc{o}_t$, the agent takes an action $\vc{a}_t \in \mathcal{A}$, where the discrete action space $\mathcal{A}$ is defined as \{MoveForward, TurnLeft, TurnRight, Stop\}, which is the default setting in the Habitat~\cite{habitat19iccv} environment.
The episode is considered successful only if the agent calls the 'Stop' action at the location close enough to the pre-defined target.
Otherwise, the episode is considered failed, and the agent does not receive a success reward.
% Otherwise, if the agent calls 'STOP' action far from the target location or does not call 'STOP' action within the episode length, the episode is unsuccessful.

% Define a reward function
\myparagraph{Reward function.}
Our reward function is designed to encourage the agent to reach the goal, which is slightly adapted from the default setting in the Habitat:
\begin{align*}
r_t(\vc{s}_t,\vc{a}_t) =\begin{cases} r_{s}, \hspace{2.3cm}\text{if} \; d^{g}_t < 0.1\text{ and }\vc{a}_t = \text{Stop} \\
r_{d} + w_1(d^{g}_{t-1} - d^{g}_{t}) - w_2(d^{s}_{t-1} - d^{s}_{t}),\; \text{others}
\end{cases} \label{reward_function} 
\end{align*}
where $r_{s}$ is the reward when the episode ends successfully, $r_{d}$ is a fixed negative live penalty, and $d^{g}_t$ and $d^{s}_t$ are the distances from the agent to the goal and start point at time $t$. 
The term $ d^{s}_{t-1} - d^{s}_{t}$ is designed to encourage exploration at the beginning of episode to prevent the agent from being trapped in a narrow room.
We set $r_{s}$, $r_{d}$, $w_1$, and $w_2$ as 10, -0.1, 1 and 0.1 for all the experiments.
% \maks{I am not sure how $d^{s}_t$ - $d^{s}_{t-1}$ encourages exploration? you can do step fwd and bwd and still maximize this term without exploring. are we actually using $d^{s}_t$ $d^{s}_{0}$ instead?}
% \maks{also, I think there is a typo in the last sentence: should the $r_{d}$ be -0.1? and $r_{s}$ be 10? }
% \sehoon{I update a bit. Qian, check the numbers.}
% \qian{Checked. It's correct}

\subsection{Navigation baseline} \label{sec:baseline}

\begin{figure}
    \centering
    \includegraphics[width=1.0\linewidth]{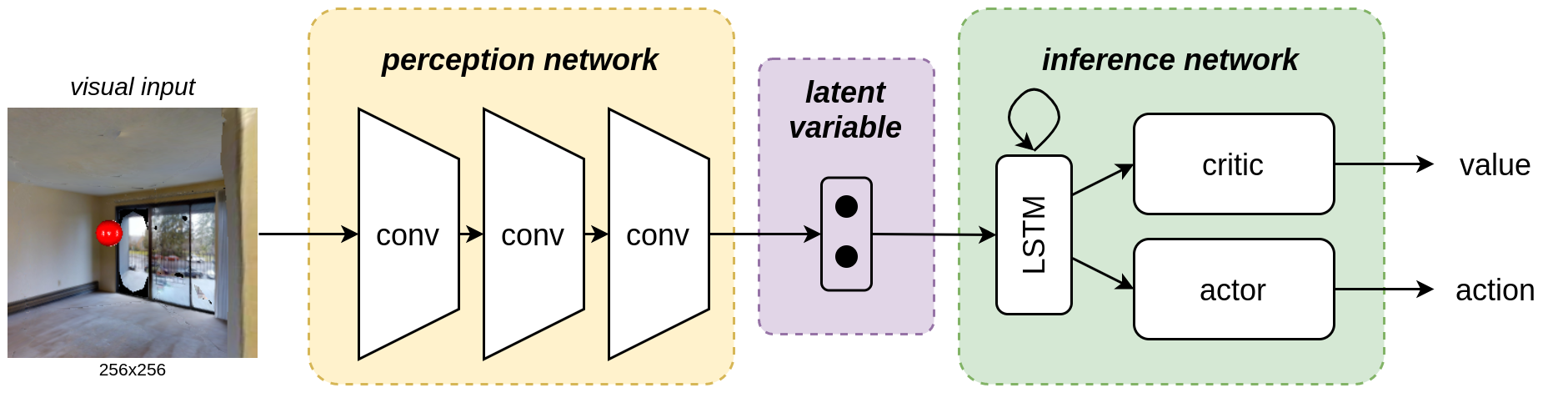}
    \caption{A network architecture of a navigation agent, which consists of the perception and inference networks.}
    \label{fig:nav_baseline}
\end{figure}

% \sehoon{Move to baseline}
% Now suppose we have trained a clever agent with a good policy $\pi$, which could successfully find a target under certain sensor configuration.
% We hope to apply the navigation skills learned by training agent to the new tasks(sensor changing, policy changing), where simply using the old policy would result in bad performance. 
% There are two main parts in our method: Firstly, we will train a clever agent with a certain sensor configuration to find a given target. 
% Secondly, we will figure out an efficient learning framework for the new agent to quickly learn the navigation skills of the trained agent with low cost of data collection. 
As the first step, we train a baseline navigation agent that successfully finds an object based on visual inputs.
Our algorithm particularly requires a highly effective agent because it defines the nominal behaviors of all the agents.
We train a baseline agent with a \emph{standard} configuration with the $1.25$~m camera height with a red color target.
% Please note that we can train a baseline agent in many different ways: for instance, we can train an effective agent by first training a teacher agent with the privileged information, such as global position and orientation information, as described in the ``learning by cheating’’ paper~\cite{chen2020learning}.
% However, the policy obtained with this approach may not be successfully transferred to the agents with visual inputs due to potential degenerate cases.

We use an end-to-end RL framework to train an agent to find a specified target with a single observation function.
Our policy architecture consists of two components: the perception network and the inference network (Figure~\ref{fig:nav_baseline}).
For the perception network, we use a simple 3-layer CNN visual encoder to extract features, or the latent variables, of the RGB image input.
The inference network processes the features through a Long Short-Term Memory network (LSTM)~\cite{hochreiter1997long} and computes the final actions and values.
We optimize the entire policy network using Proximal Policy Optimization (PPO) algorithm~\cite{schulman2017proximal}.
% In our experiments, the learned baseline agent achieves near $100$\% success rates for the given navigation tasks.
% We then use Proximal Policy Optimization(PPO) algorithm to process the output hidden layer.
% The hidden states are passed through by critic neural network to output values and actor network to output actions, respectively. 
% From our experiment, the training is quite successful and the trained agent could accomplish 100$\%$ of the navigation tasks in our validation set. 

\subsection{Adaptation with Supervised Learning} \label{sec:supervised}

\begin{figure}
    \centering
    \includegraphics[width=1.0\linewidth]{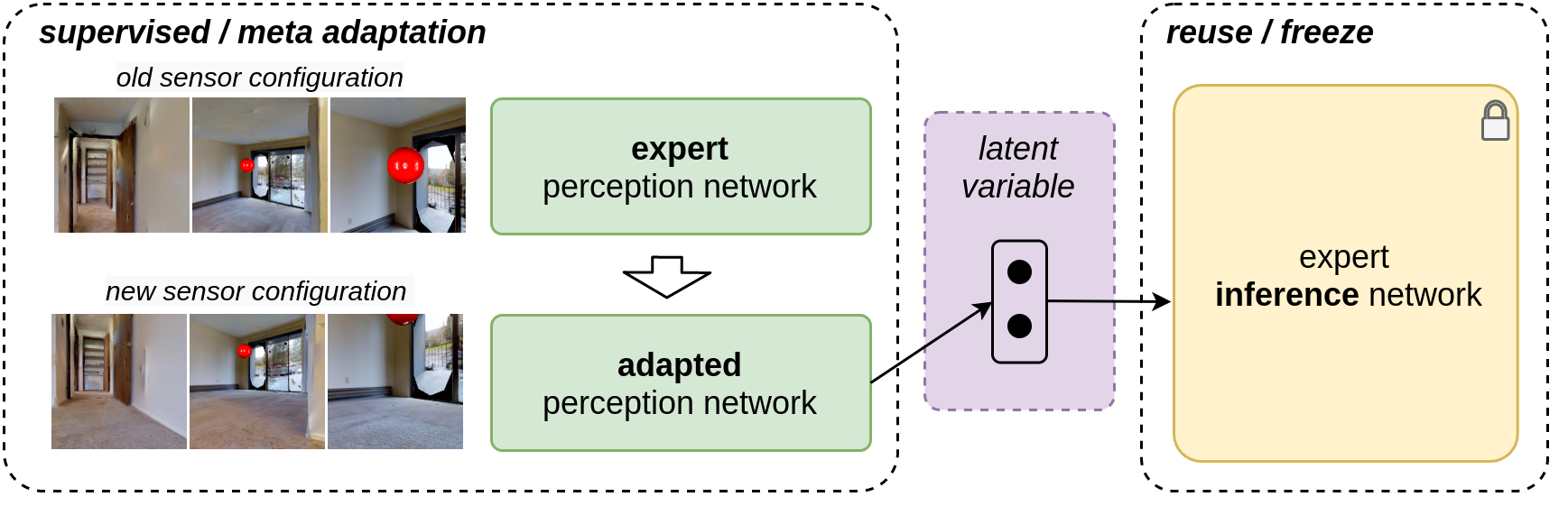} 
    % {figures/adapt_train.png}
    % \includegraphics[width=1.0\linewidth]{figures/latent_SL.png}
    \caption{Adaptation via supervised learning or meta learning. We only adapt the perception network while freezing the inference network.}
    \label{fig:latent_sl}
\end{figure}

Once we obtain a baseline agent, the next step is to adapt it to an updated observation function $\vc{o}_t \sim \tilde{c}(\vc{s}_t)$ that is different from the original function $c$.
A naive solution is to train a new policy from scratch; however, this approach takes a lot of computational time and does not guarantee that the same hyperparameters can learn a satisfactory policy in a new scenario.
Alternatively, a more structured approach is domain randomization (DR) ~\cite{tobin2017domain,sadeghi2016cad2rl}, where we train a single navigation agent while randomizing observation functions during the training process.
However, DR can learn a conservative policy that is suboptimal for many tasks.
% In our experience, domain randomization can take very long time for convergence and sometime show the poor performance at the end. \maks{Qian: will we have domain randomization with the same compute budget results to support this claim?}  \sehoon{I tone down this sentence, just in case.} \qian{from the results, DR will not take more time. The success rate and spl are just similar as SL method}

Instead, we propose only to adapt the perception network to a new observation function $\tilde{c}$ using supervised learning (Figure~\ref{fig:latent_sl}), which is similar to the idea of the prior work of Gordon et al.~\cite{gordon2019splitnet}.
More specifically, we adapt the learned perception network to produce the same latent variables from the collected images, which will be the inputs to the inference network.
We do not change the inference network because the agent can use the same actions to solve the problem if the current state is inferred correctly.
% (which remains unchanged since the navigation baseline training phase in Section~\ref{sec:baseline}).

For all the states $\{\vc{s}_1, \cdots, \vc{s}_K\}$ in the sampled trajectories, we collect all the tuples of $(c(\vc{s}_t), \tilde{c}(\vc{s}_t), \phi(\vc{o}_t))$ where $\phi$ is the perception network that outputs the latent state from the visual observations.
The same simulator state $s_{t}$ is used as input to both $c$ and $\tilde{c}$, and the difference in $o_{t}$ is reflected in the sensory observation (e.g. camera view from a different sensor heights or a distinct target object).
Then we can train a new perception function $\tilde{\phi}(\vc{s}_t)$ by minimizing the loss $\mathcal{L} = \sum |\phi(c(\vc{s}_t)) - \tilde{\phi}(\tilde{c}(\vc{s}_t))|$.
Our preliminary experiment indicates that this adaptation with SL works well for a target navigation task in various scenarios given enough data from the new environment.

\subsection{Adaptation with Meta-Learning} \label{sec:meta}
\begin{algorithm}
    \caption{ A Few Shot Adaptation with Meta-Learning  }
        \begin{algorithmic}[1]
        \STATE Initialize task distribution $p(\mathcal{T})$, learning rates $\alpha$ and $\beta$, a number of gradient steps $K$, and model parameter $\theta$\\
        \WHILE{not done}
            \STATE{Sample batch of tasks $\mathcal{T}_i$ based on $p(\mathcal{T})$}
            \FORALL{$\mathcal{T}_i$}
                 \STATE{Calculate $\nabla_\theta\mathcal{L}_{\mathcal{T}_i}(f_\theta)$ and compute adapted parameters with $K$ gradient descent: $\theta'_i = \theta - \alpha \nabla_\theta\mathcal{L}_{\mathcal{T}_i}(f_\theta)$}\\
            \ENDFOR
            \STATE{$\theta \leftarrow \theta - \beta \nabla_{\theta} \sum_{\mathcal{T}_i \sim p(\mathcal{T})} \mathcal{L}_{\mathcal{T}_i} (f_{\theta'_i})$}
        \ENDWHILE
        \end{algorithmic}

\label{alg:maml}
\end{algorithm}

However, if we aim to apply the SL approach to real-world scenarios, it is likely to be infeasible because it requires a lot of samples from the target-environment.
In addition, it requires us to collect the observation $\tilde{c}(\vc{s}_t)$ from the same simulation trajectory $\{\vc{s}_1, \cdots, \vc{s}_K\}$.
But it is difficult for a robot to exactly track the given simulation trajectory in the real-world environment due to the difference in observation functions and control errors, unless we exploit other frameworks, such as Simultaneous Localization and Mapping (SLAM)~\cite{bailey2006simultaneous} or external motion capture systems.

Instead, it will be more data-efficient if we can adapt to a new scenario using only a handful of examples via meta-learning.
In our scenario, this corresponds to collection of observations from a few key places, such as the four corners of the room or locations near the goal object.
Please note that we simplified the problem by assuming that the sim-to-real gap is only introduced by a new observation function, while the other sim-to-real factors have minimal impacts due to the photo-realistic rendering of the environment.

% \maks{our approach also requires a lot of the same states during the training phase, do we assume that sim2real has no gap? like generate those states in the simulation and when we transfer we only need "few" real states?}
% \maks{perhaps we should be a little more specific about how the training data is collected for meta-training}

We propose a data-efficient adaptation via meta-learning, which requires only a few shots.
We exploit the existing meta-learning framework, Model Agnostic Meta Learning~\cite{finn2017model}, which explicitly learns to make a quick adaptation to the perception network $\tilde{\phi}$ with a small number of gradient steps.
In our scenario, we consider a distribution of tasks, $p(\mathcal{T})$, where each task $\mathcal{T}_i$ is to navigate with the given observation function $\tilde{c}_i$.
We train our meta-perception network $\tilde{\phi}_{\theta}$, which is parameterized by $\theta$,  to learn an unseen task from only $M\leq10$ samples.
To this end, we first sample a batch of $N$ tasks $\mathcal{T}_{1 \cdots N} $ from the distribution and compute the adapted parameters $\theta_{1 \cdots N}$ by taking $K$ gradient steps.
Then we update the meta-parameters $\theta$ by aggregating the changes of the adapted parameters,  $\theta_{1 \cdots N}$.
The algorithm is summarized in Algorithm~\ref{alg:maml}. 
For more details, please refer to the original paper~\cite{finn2017model}.

% By supervised learning, we could make the agent adapt to new tasks. 
% However, we can not regard supervised learning as a generalized method, since it’s largely limited to its training dataset and it still requires a large amount of data in a new task. 
% To learn a more generalized visual encoder, we apply Model-Agnostic Meta-Learning to enable the agent to learn based on few shots of new task.

% The MAML algorithm optimizes for fast adaptation to new tasks. If the distribution of training and testing tasks are sufficiently similar then a network trained with MAML should quickly adapt to novel test tasks.

% Typically, if we hope the agent could quickly learn to navigate to a brand new target, it would be hard to create dataset based on this target.
% All we get are just a few images containing the target.
% So in this case, we have to learn a generalized visual encoder which could quickly adapt to the new target.

%% file: Experiment.tex
\section{EXPERIMENTAL EVALUATION}

In this section, we discuss the simulation experiments for validating the proposed algorithm.
We design the experiments to answer the following questions.
\begin{enumerate}
    \item Can our algorithm effectively adapt to unseen observation functions?
    \item Does our meta-adaptation show better generalization than other algorithms?
    \item How does the selection of hyper-parameters impact the performance?
\end{enumerate}

\subsection{Experiment Setup}
\label{exp_set}
\begin{figure*}
    \centering
    \includegraphics[width=1.0\linewidth]{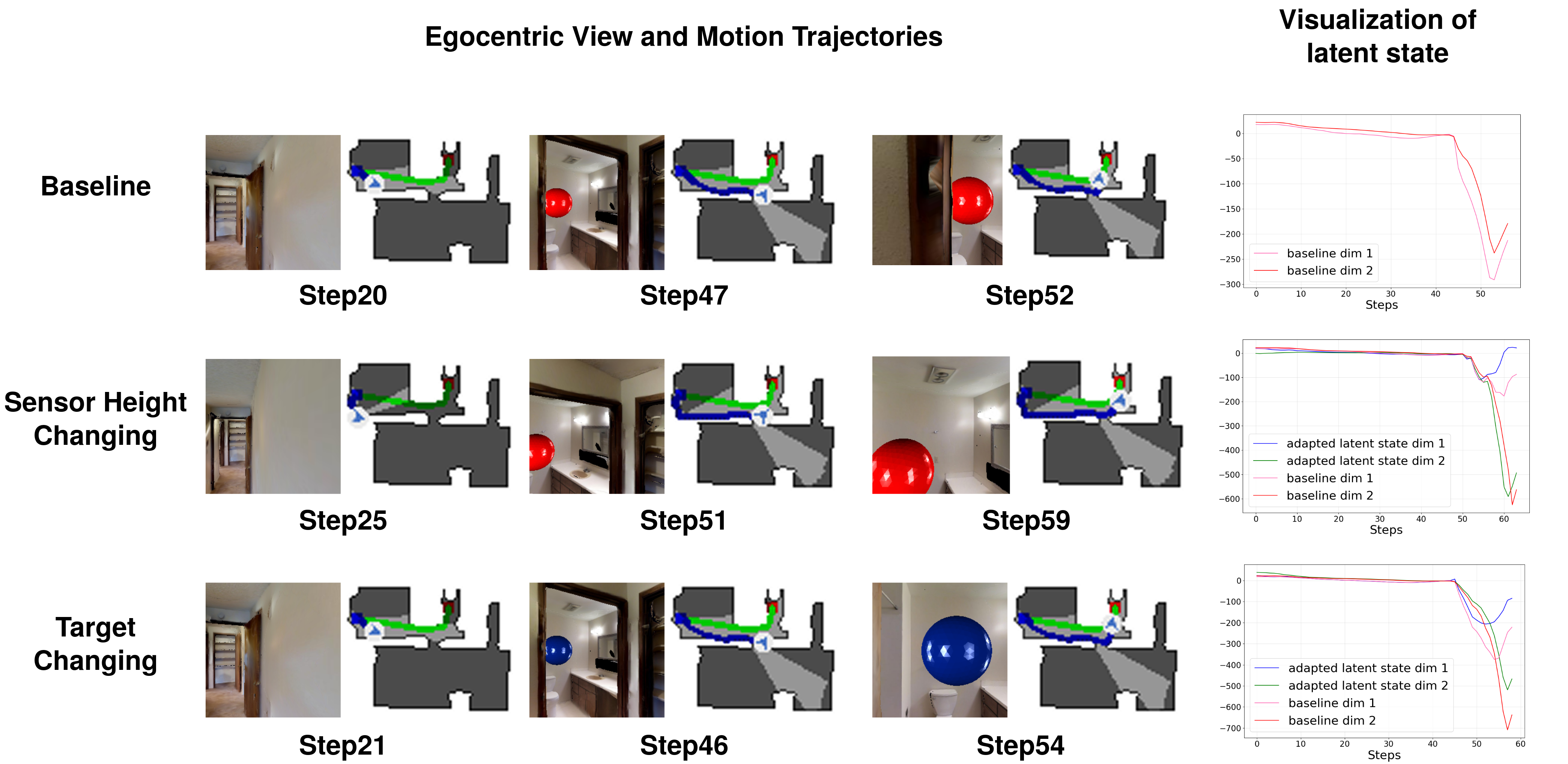}
    \caption{Learned Behaviors of the baseline agent with the default setting (\textbf{Top}) and the adapted agent with meta-learning for the camera height change (\textbf{Middle}) and the target object change (\textbf{Bottom}). We show three observations and three map images, which are sampled at different moments. For all experiments, green lines indicate the shortest path between the starting and goal locations. The rightmost column represents the trajectories of two dimensional latent variables over time.
    % \sehoon{This figure can be improved a lot. Basically, it is too sparse, the heights are not matched. See if 1) Delete the first column (name), 2) Make the center images larger, and 3) the latent space image smaller.}
    % \sehoon{Stage 1, 2, 3 does not mean anything. Can you specify their time? If they are not consistant, you can just have one merged header ``Motion Trajectories''.}
    % \sehoon{Did we use one dimensional latent space? I think it is better to show two dimensiol plots.}
    % \sehoon{Also put vertical dotted lines to the latent plot so that we can identify their moment.}
    }
    \label{fig:motion}
\end{figure*}

We use the Habitat simulator as our testing platform, which offers photo-realistic rendering at high speed, up to over several thousand FPS on a single GPU.
All the experiments are conducted on a laptop equipped with an 8GB Intel i7 8750H Processor and a GTX1060 GPU.
We firstly train the navigation baseline (Section~\ref{sec:baseline}) in the Habitat Avonia environment to find a red ball with the 1.25m camera height.
The training dataset contains 4000 episodes, and the validation set contains 200 new episodes with randomized initial agent positions and target object locations.
% \sehoon{Qian, what is distinct initialization? random initialization?} \qian{Yes, the starting/ending position/location are initialized randomly, and different from each other}

We learn policies using PPO~\cite{schulman2017proximal}, where the discount factor is set to 0.99, and the GAE parameter is 0.95.
After 6 hours of training, the agent could navigate to the target with 84$\%$ success rate and 0.69 success weighted path length (SPL)~\cite{anderson2018evaluation}.
For adaptation wiht meta learning, our MAML implementation uses the same inner learning rate of 2e-4 and outer learning rate of 2e-4, three-shots(all containing the object), and ten gradient step updates.
% The size of the query set \revised{(used for outer loop updates of MAML)} is 50, where 25\% to 50\% of the images contain the target object.
% I deleted the above sentence because we did not mention it in the algorithm section.

% For both experiments, we use the same inner learning rate(2e-4) and outer learning rate(2e-4), and implement MAML with 3-shots(all containing the object), 10-steps updates.
% The size of the query sets are 50 in both cases, and we make sure that the number of images containing the target take up 1/4 to 1/2 of the query set.
% During training, we will visualize the losses of the latent values before and after inner updates every ten outer update steps, in both training and testing sets.
% Each losses are calculated on average of ten trails.
% After the losses of both training and testing converge, we test the SPL and success rate of both seen and unseen observation functions in the validation set of our navigation baseline. 

\subsection{Adaptation Experiments with Meta-Learning}
\label{sec:exp}

Figure~\ref{fig:motion} shows the rollouts of three agents: \begin{enumerate}
    \item \textbf{Navigation Baseline:} The agent is trained and 
    evaluated with fixed sensor (1.25m) and target configuration.
    \item \textbf{Sensor Height Changing:} The agent is meta-trained with the heights between 0.6m and 1.3m and evaluated at 1.8m after meta-adaptation.
    \item \textbf{Target Changing:} The agent is meta-trained with 10 target colors and evaluated with a new dark-blue color after meta-adaptation.
\end{enumerate}

Each row contains three egocentric observations and three map images collected at three different moments: when an agent is searching the ball, locating the ball, and getting close to the ball, respectively.
As shown in Figure~\ref{fig:motion}, our algorithm successfully adapts the policy to new scenarios using only three-shots.
As expected, the adapted agents show very similar behaviors to the baseline agent because we only adapt the perception network.
The rightmost column of Figure~\ref{fig:motion} plots two-dimensional latent values in the rollouts, which become larger when the agent detects the target object.
Meta-adapted agents in both scenarios closely track the baseline agent's latent features, 
% except when it is in near proximity to the target object.
% We suspect that this is because there do not exist many samples when the agent is close to the target.
% However, the adapted agents 
and are able to call the ``STOP'' action to end the episode successfully.

\begin{figure}
    \centering
    \includegraphics[width=0.48\linewidth]{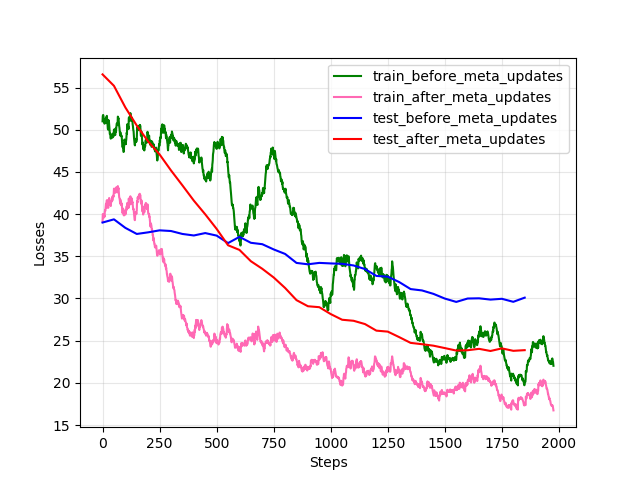}
    \includegraphics[width=0.48\linewidth]{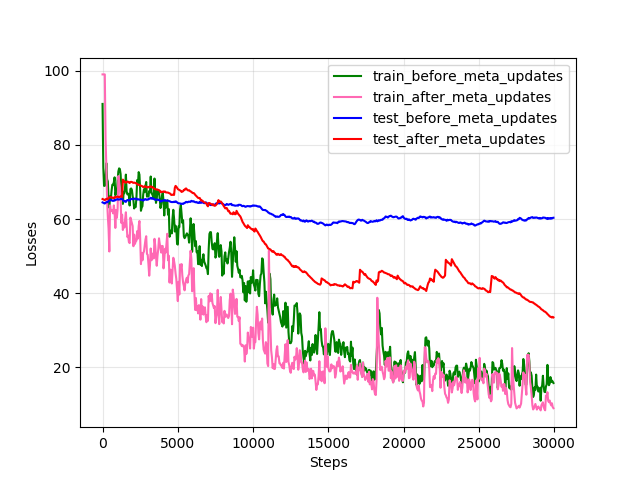}
    \caption{Learning curves of meta-adaptation for two tasks: sensor height changes (\textbf{Left}) and object color changes (\textbf{Right}). The results indicate that meta learning can successfully adapt to new scenarios (blue to red).}
    \label{fig:meta_results}
\end{figure}

Figure~\ref{fig:meta_results} illustrates the learning curves of two meta-adapted agents.
The training takes only 15 minutes for the sensor height changing scenario while taking three hours for the color-changing problem.
For both scenarios, the testing loss after meta-updates is much lower than the testing loss before beta-updates, highlighting the importance of the meta-updates in our algorithm.
And we also want to note that the testing loss for the color-changing problem is still going down even when the training curve is saturated.

We compare the proposed algorithm against the navigation baseline and a few other training methods:
\begin{enumerate}
    \item \textbf{Navigation Baseline:} The agent is trained with a nominal observation function as described in Section~\ref{sec:baseline}.
    \item \textbf{Domain Randomization:} The agent is trained with randomized observation functions within the training range. 
    % \item \textbf{Domain Randomization with target embedding:} This is an amended version of Domain Randomization, where the input is augmented by an observation of the target. Similar to \cite{zhu2016targetdriven}, the visual encoder is weight-shared for processing current and target observations.
    % In this case, such augmentation serves as an oversampling mechanism, balancing the visible/non-visible target image distribution for perception network training.
    \item \textbf{Domain Randomization with target embedding:} Inspired by the work of Zhu et al.~\cite{zhu2017target}, we amend domain randomization to take an observation of the target as additional inputs.
    Such augmentation serves as an oversampling mechanism that balances the distribution of target and non-target images.
    \item \textbf{Supervised Learning (SL):} The agent is adapted with supervised learning as described in Section~\ref{sec:supervised}.
    \item \textbf{SL + Few Shots:} Based on the perception network learned by SL, we further refine the model by leveraging three images from the target environment, which uses the same amount of data as meta-learning.
    \item \textbf{MAML + Few Shot Adaptation} The agent is trained using the method described in Section~\ref{sec:meta} and task settings described in Section~\ref{sec:exp}
\end{enumerate}

We apply these methods to two tasks: camera sensor height changes task and target color changes task. In both tasks, the training and testing range of either method are the same. 
For all experiments, we train each policy until they are fully converged.
We did not set the same computation budgets because some methods (Navigation Baseline, Domain Randomization) are pure reinforcement learning and the others combine reinforcement learning and supervised learning. We run 200 episodes for each agent and take the average SPL and success rate to ensure the accuracy.

% (1)\textbf{Navigation Baseline:}
% It is trained with a single observation function as described in Section~\ref{sec:baseline}.
% (2)\textbf{Domain Randomization:}
% The agent is trained with randomized observation functions. 
% In sensor height changing case, height of the agent is initialized uniformly between 0.6m to 1.6m. 
% In color changing case, the color of the object is initialized randomly based on 8 classes of colors described in Section~\ref{exp_set}. 
% (3)\textbf{Domain Randomization with target embeddings:}
% It's a enhanced version of Domain Randomization inspired by Zhu~\cite{zhu2016targetdriven}, where a target image processed by a visual encoder of same architecture is added to the input.
% The target image is either the observation of red ball in certain sensor height(height changing), or the observation of a different color of ball(color changing).
% (4)\textbf{SL(Supervised Learning):}
% Using the method described in Section~\ref{sec:supervised}.
% The training and testing set are the same as described in Section~\ref{exp_set}.
% (5)\textbf{SL + Few Shot Adaptation:}
% Based on the visual encoder learned by (4), we conduct 3 shots 10 steps update to adapt to new scenes.
% (6)\textbf{MAML + Few Shot Adaptation:}
% Using the method described in Section~\ref{sec:meta}, and the training settings described in Section~\ref{exp_set}.
% \maks{maybe we should be somewhat transparent here and add an appendix table with how much compute / data was used by each method.}

\subsection{Comparison of Adaptation Algorithms}

\begin{table} 
\begin{center}
\caption{\label{tab:results}Comparison of Adaptation Algorithms}   
\begin{tabular}{lllll}    
\toprule & \multicolumn{2}{c}{Training} & \multicolumn{2}{c}{Testing} \\
 \cmidrule {2-5} \textbf{Task:} Camera Height Changes & SPL & Success & SPL & Success\\    \midrule  
 Navigation Baseline & 0.69 & 0.84 & 0.00 & 0.00\\  
 Domain Randomization & \textbf{0.71} & \textbf{0.86} & 0.04 & 0.10\\
 DR + target embedding & 0.69 & 0.82 & 0.06 & 0.09 \\
 SL & 0.66 & 0.83 & 0.02 & 0.05 \\  
 SL + Few Shots & 0.61 & 0.75 & 0.01 & 0.03 \\ 
 \textbf{MAML + Few Shot Adaptation} & 0.67 & 0.84 & \textbf{0.18} & \textbf{0.25} \\
 \bottomrule   
\toprule& \multicolumn{2}{c}{Training} & \multicolumn{2}{c}{Testing} \\
 \cmidrule {2 -5} \textbf{Task:} Target Color Changes & SPL & Success & SPL & Success\\    \midrule  
 Navigation Baseline & \textbf{0.69} & \textbf{0.84} & 0.06 & 0.10\\  
 Domain Randomization & 0.41 & 0.50 & 0.16 & 0.21 \\ 
 DR + target embedding & 0.43 & 0.52 & 0.16 & 0.22 \\ 
 SL & 0.63 & 0.82 & 0.18 & 0.24 \\ 
 SL + Few Shots & 0.60 & 0.78 & 0.08 & 0.11 \\ 
 \textbf{MAML + Few Shot Adaptation} & 0.63 & 0.80 & \textbf{0.50} & \textbf{0.70}\\
 \bottomrule    
\end{tabular}  

\end{center}
\end{table}

The results are summarized in Table~\ref{tab:results}.
During the training, all of the methods produce approximately equal results, and all perform on par with the Navigation Baseline.
However, in the testing setting our algorithm, adaption with meta-learning shows the best performance in both scenarios, while others approaches perform poorly on the unseen tasks.
In the color changing case, our algorithm dramatically outperforms other methods, and the SPL and success rate are close to that of baseline agent in the training agent.

\subsection{Analysis}

% \qian{I think the 'ablation study' should be discussing the removal of meta updates during testing. There are four method to try: 1. SL 2.SL + few shots updates based on the few shots 3. MAML + without few shots update 4. MAML}

\begin{figure}
    \centering
    \includegraphics[width=0.48\linewidth]{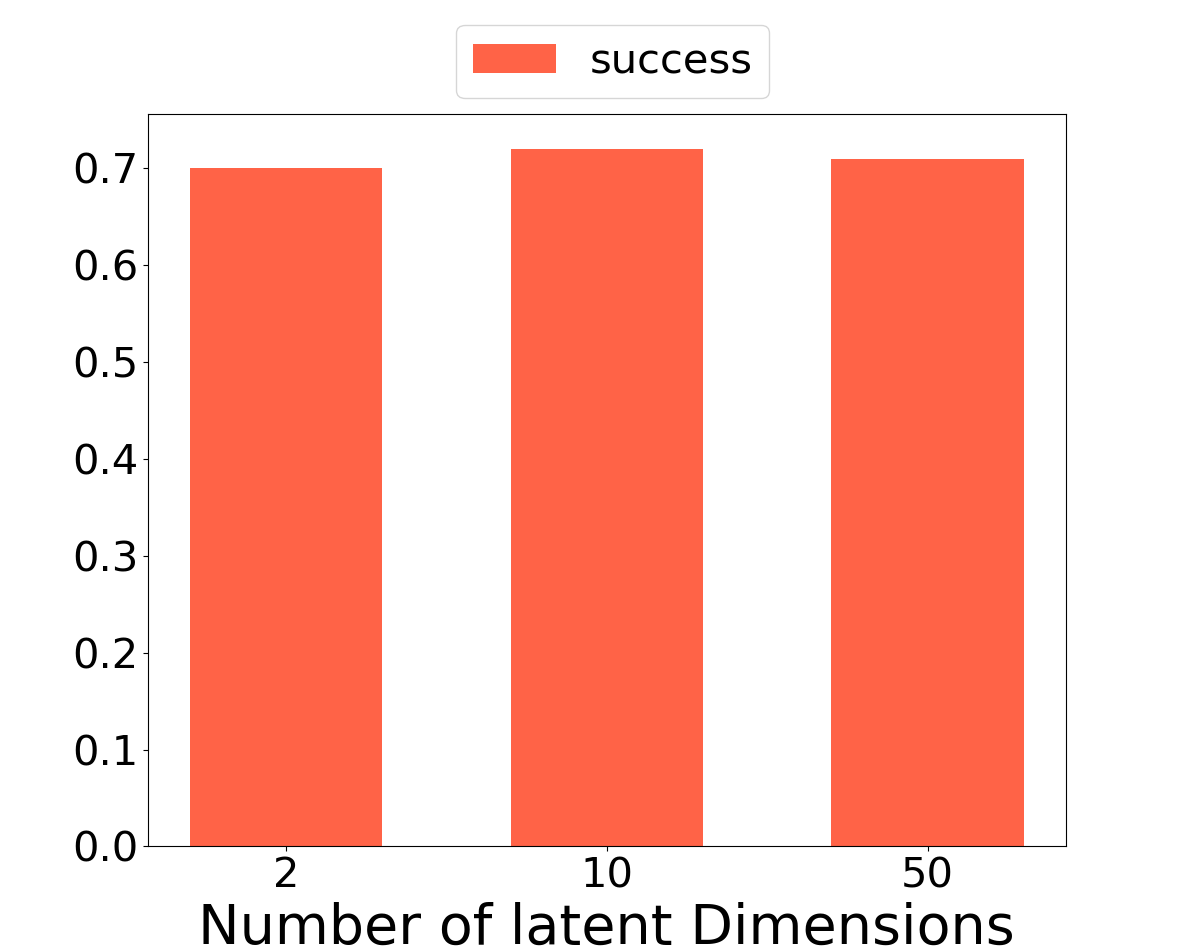}
    \includegraphics[width=0.48\linewidth]{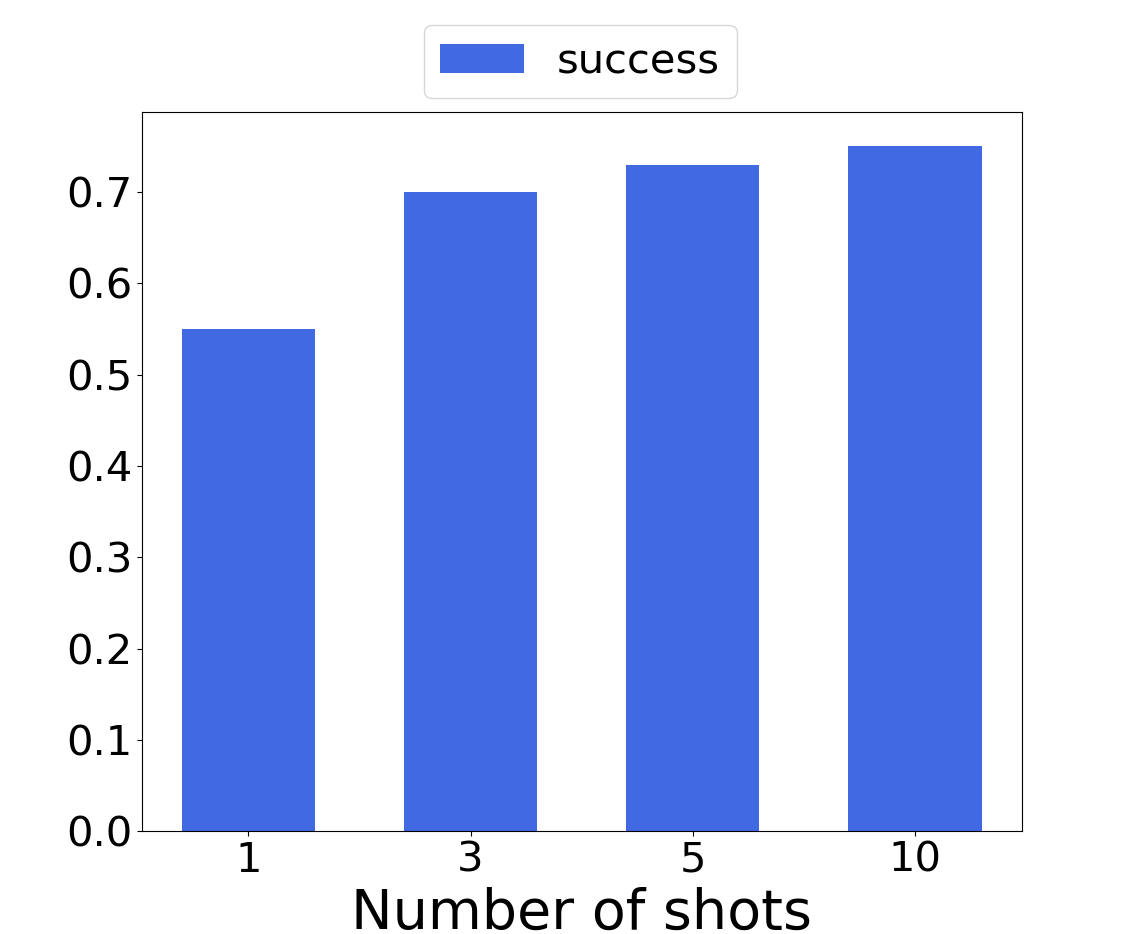}
    \caption{Analysis on meta-learning adaptation. We report the average success rates that are measured in both sensor height changing and target color changing scenarios. We investigate the number of the latent dimension (\textbf{Left}) and the number of shots (\textbf{Right}). More shots lead to better success rates but the number of latent dimensions do not have significant impacts on the success rate.
    }
    \label{fig:ablation}
\end{figure}

We further analyze the proposed algorithm by conducting an ablation study.
Particularly, we investigate the impact of two hyperparameters: the dimension of the latent space between the perception and inference networks, which is studied based on three-shot MAML adaptation, and the number of shots in meta-adaptation studied with two-dimensional latent space.
Intuitively, the dimension of the latent space is expected to have an impact on the performance: a smaller dimension may be easier to adapt but also may result in a more convoluted latent space.
We also expect that more shots grant the meta-learning process more adaptation capability while making the entire network harder to learn.

We compare the performance of meta-learning with different hyperparameters in Figure~\ref{fig:ablation}. 
% The impact of the latent dimension size is studied with three-shot MAML adaptation, and the effect of the number of shots is studied with two-dimensional latent space.
All of the experiments are evaluated in the testing sets for both sensor height changing and object color-changing tasks.
Against our intuition, the dimension of the latent space does not significantly affect the success rate.
We suspect that this is because a low-dimensional embedding is sufficient for our testing scenarios: please note that an agent can achieve the 70\% success rate with only two-dimensional latent features.
Then a high dimensional space is expected to have a similar embedding structure, which is also easy to learn.
%or all experiments, we choose two dimensional latent space for analysis.
On the other hand, the number of shots positively impacts the success rate: one-shot adaptation results in only a 55\% success rate while the ten-shot adaptation achieves a 73\% success rate.
However, more shots require us to collect more real-world data, which can be cumbersome on real robots.
We choose three shots for all the experiments, which shows near the optimal performance with the minimal number of additional examples.

% In our first experiment, we use supervised learning and domain randomization method to adapt to new sensor height observations. The baseline navigation is trained at 1.25m sensor height, with 1.4m fixed ball height. Then we apply our supervised learning method to learn the latent space based on 2 different heights: 0.5m and 2.0m, and test the SPL and success rate on these two types of observations. In these two cases, the old scene is 1.25m sensor height, and the new scenes are 0.5m and 2.0m sensor heights.

% Also we test the results of domain randomization, where random sensor heights of sensors ranging from 0.6 to 1.8m are initialized during RL training. We test the result of old scene(0.6-1.8m), and new scene(0.5m and 2.0m).

% % \maks{what is the number of rollouts used to generate the results in the table?}

% The second is the object changing experiment. The training set contains ten different colors of balls, and the testing set contains three different colors of balls.

% Baseline: trained only in the red ball case, tested in the testing set. 
% Domain Randomization: trained randomly in the training set(ten colors), tested in the testing set. 
% Yuke Zhu's method: use the method described in \cite{zhu2016targetdriven}, where the target image is added to the input of visual encoder. 
% Supervised Learning: learn the latent space in the training set using supervised learning
% Meta+few shots update: learn the latent space in the training set using MAML

%% file: Conclusion.tex
\section{CONCLUSION}

We showed that meta-learning on an embedding space produced by a visual encoder demonstrates promising results at the adaptation of the navigation agents in the new sensory configurations. 
We effectively apply our method to two different adaptation scenarios, object color and sensor height changing problems, in the target-driven visual navigation task. 
Our method requires only a handful of shots from the target environments and outperforms the baseline generalization methods described in the paper.
% , such as domain randomization or the adaptation method of Zhu et al.~\cite{zhu2017target}.

When we consider the fact that only a few shots are required at the adaptation stage, our method could potentially be effective in the sim-to-real transfer.
In this approach, the agent will be firstly trained in the simulated environment utilizing the scalability of the recent deep RL algorithms, without the safety risk to the robot or its surroundings.
Then the learned policy can be adapted to the real-world scenario via the proposed meta-adaptation algorithm, which requires only a few real-world images.
We plan to further investigate our algorithm on real robots, such as Fetch~\cite{wise2016fetch} or TurtleBot~\cite{turtlebot}.

%  Future work
While we evaluate our algorithm on two different scenarios, there exists a variety of sensory adaptation scenarios that can be further tested in the future.
Such variations include environment lighting, camera properties (FoV, lens distortions), multiple cameras, different sensors (LIDAR and IMUs), more complex object shapes, and the combination of all of these factors.
% We also plan to investigate the adaptation of the inference network, such as adapting the actions from a wheeled robot to a legged robot, which will complete the transfer of the learned policy between heterogeneous robots.